\documentclass{article}

\pdfoutput=1 
 
\usepackage{arxiv}

\usepackage[utf8]{inputenc} 
\usepackage[T1]{fontenc}    
\usepackage{hyperref}       
\usepackage{url}            
\usepackage{booktabs}       
\usepackage{amsfonts}       
\usepackage{nicefrac}       
\usepackage{microtype}      
\usepackage{lipsum}		
\usepackage{graphicx}
\usepackage[numbers]{natbib}
\usepackage{doi}
\usepackage{amsmath}

\usepackage{algorithmic}
\usepackage[ruled,vlined]{algorithm2e}
\usepackage{placeins}

\usepackage[caption=false,font=footnotesize]{subfig}

\usepackage{multirow}

\title{Data Expansion using Back Translation and Paraphrasing for Hate Speech Detection}

\author{Djamila Romaissa Beddiar and Md Saroar Jahan and Mourad Oussalah\\
	Center for Machine Vision and Signal Analysis, \\
	University of Oulu, \\
	Finland\\
	\texttt{Djamila.Beddiar@oulu.fi} \\

}

\date{}

\renewcommand{\headeright}{Under Review}

\begin{document}
\maketitle

\begin{abstract}
With proliferation of user generated contents in social media platforms, establishing mechanisms to automatically identify toxic and abusive content becomes a prime concern for regulators, researchers, and society. Keeping the balance between freedom of speech and respecting each other dignity is a major concern of social media platform regulators. Although, automatic detection of offensive content using deep learning approaches seems to provide encouraging results, training deep learning-based models requires large amounts of high-quality labeled data, which is often missing. In this regard, we present in this paper a new deep learning-based method that fuses a Back Translation method, and a Paraphrasing technique for data augmentation. Our pipeline investigates different word-embedding-based architectures for classification of hate speech. The back translation technique relies on an encoder-decoder architecture pre-trained on a large corpus and mostly used for machine translation. In addition, paraphrasing exploits the transformer model and the mixture of experts to generate diverse paraphrases. Finally, LSTM, and CNN are compared to seek enhanced classification results. We evaluate our proposal on five publicly available datasets; namely, AskFm corpus, Formspring dataset, Warner and Waseem dataset, Olid, and Wikipedia toxic comments dataset. The performance of the proposal together with comparison to some related state-of-art results demonstrate the effectiveness and soundness of our proposal. 
\end{abstract}

\keywords{Cyberbullying Detection \and Hate Speech \and Back Translation \and Paraphrasing \and NLP Transformers \and Encoder-decoder.}

\section{Introduction}
\begin{figure}
\includegraphics[width=\textwidth, height = 10cm]{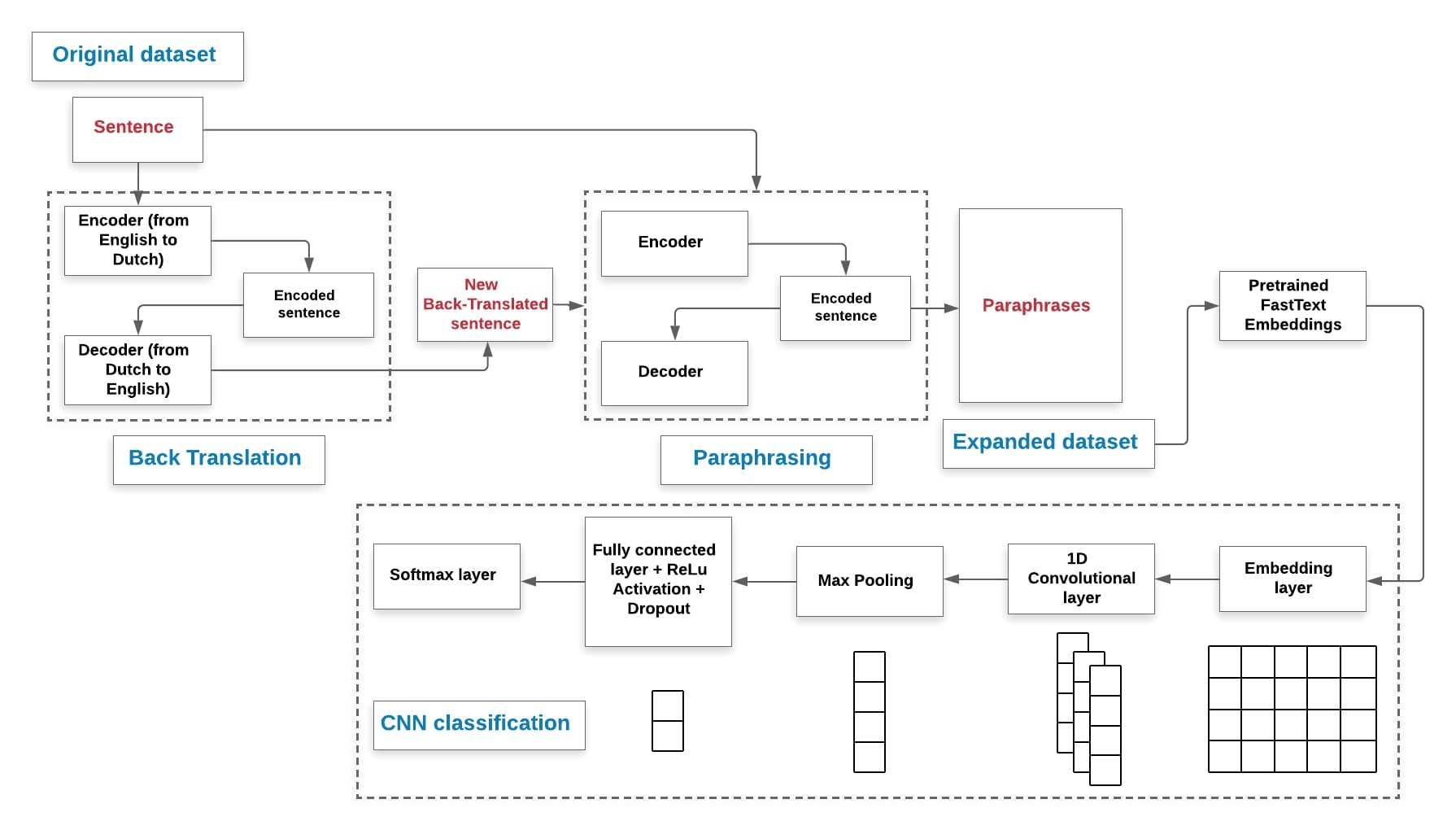}
\caption{General scheme of our proposal where expanded dataset is generated from original dataset using back translation and paraphrasing. The FastText embeddings are then calculated before classifying hate speech using CNN model. In this figure we illustrate the architecture of the CNN network.} 
\label{Fig:general}
\end{figure}

With the exponential increase in use of social media platforms where people can freely communicate their opinions and thoughts, online hate speech has seen the emergence of appropriate ecosystem, which caused concerns to authority, researchers, and society, especially in the last decade. Indeed, the easy access to various social media platforms as well as the anonymization schemes have boosted the scope of offensive online content and harassment cases. In addition, some political and racist organizations have exploited such channels to propagate toxic content such as hate speeches \cite{cao2020deephate}. Hate speech is defined in the Cambridge dictionary as: "public speech that expresses hate or encourages violence towards a person or group based on something such as race, religion, sex, or sexual orientation" \cite{Cambridge}. Maintaining a balance between freedom of speech and societal protection needs while respecting cultural and gender diversity is often challenging. Especially, hate speech can lead to serious consequences on both individual and community scale, which may trigger violence and raise public safety concerns. One predominant kind of online hate speech is cyberbullying. The latter refers to intentional, cruel, and repeated behavior among peers, using electronic media and communication technologies \cite{wright2017cyberbullying}. It has been categorized into different forms that include but not limited to cyber harassment, denigration, flaming, and happy slapping. This comprises posts associated to person's threat, false information about person with the intention to cause harm and direct insult \cite{kowalski2012cyberbullying}. The act of bullying and harming people has many impacts and consequences ranging from emotional to psychological and physical states. Therefore, it is important to detect hate speech and cyberbullying in online social media to confront pervasive dangerous situations. Automatic detection of hate speech or cyberbullying requires robust, efficient and intelligent systems, which enables accurate analysis of online content. Most of existing machine learning models are very context dependent and barely migrate from one domain to another or accommodate cross-social media platforms, and often experience substantial decrease of accuracy when dealing with large amounts of data \cite{dadvar2018cyberbullying}. Another important issue that constrained the development of efficient automatic detection systems is related to the availability of high quality labeled data, which challenge the use of deep learning based approaches. Many of the existing data augmentation techniques applied in image processing and signal analysis cannot straightforwardly be replicated to textual data as they may affect the correctness of syntax and grammar or even alter the meaning of the original data. In addition, the categorical feature of words and tokens could prohibit the applications of many techniques of data augmentation such as noise injection \cite{rizos2019augment}. Due to the application of automatic hate content detection in different social media networks, it is difficult to obtain online hate speech since this is automatically deleted by the system. To deal with the issue of insufficient labeled data for hate speech detection, we present in this paper a new deep learning based method. It uses back translation and paraphrasing \cite{edunov2018understanding} techniques for data augmentation in order to ensure class balances and an embedded based deep network to yield a binary classification (hate speech versus non-hate speech case). More specifically, back translation is based on a transformer architecture that includes an encoder decoder model. The paraphrasing approach makes use of machine translation models where text is translated to an intermediate language and back to original one. In our approach, the original text is inputted to an English-French translation model, followed by a French-English mixture of experts translation model \cite{shen2019mixture}. The newly generated sentences together with the original sentences are fed to a classifier for training. For that, we use a pretrained word-embeddings as our features. We investigate the results of i) back translation based augmentation, ii) the paraphrasing based augmentation and, iii) the combination of both of them using publicly available datasets for hate speech and cyberbullying detection tasks. Fig.~\ref{Fig:general} illustrates the different steps of our proposed methodology.

The main contributions of our proposal are summarized below: 
\begin{itemize} 
    \item We performed data augmentation using a back-translation method by exploiting the pretrained English-Dutch translation model. Data from evaluated datasets is translated into Dutch and back to English to generate new augmented data.
    \item We performed data augmentation using a paraphrasing technique based on the transformer model and mixture of experts. This technique allowed us to generate varied paraphrases from a source sentence and at the same time close in meaning.
    \item Combination of both methods of data augmentation, the back-translation and the paraphrasing is also investigated to create large-scale datasets.
    \item We compared between augmenting the whole dataset or only the minority class that represents the hate class in order to identify the best strategy. 
    \item We constructed newly extended dataset by concatenating to the original data back-translation outcome, paraphrasing outcome, or both (back-propagation and paraphrasing). 
   \item We exploited FastText embeddings as the main feature representation to be fed to the classifier.
    \item We compared the results of the CNN classifier and some machine learning baseline classifiers, namely, Linear Regression and Naive Bayesian classifiers, is conveyed as well for hate speech detection.
    \item We compared between results of the CNN and the LSTM classifiers for hate speech detection.
    \item We used different evaluation metrics such as accuracy, F1 score, recall precision and ROC, in addition to an error analysis task based on human evaluation to assess and comprehend the performance of the developed approach.

\end{itemize}
The remaining structure of the paper is as follows: Section~\ref{Sec:Related work} presents related work on hate speech while focusing on cyberbullying and data augmentation. We discuss at this level some existing techniques in the literature and their limitations. Our proposed methodology including the datasets expansion and the detection mechanisms are presented in Section~\ref{Sec:Methodology}, followed by the results and discussion in Section~\ref{Sec:Results}. We conclude the paper in Section~\ref{Sec:conclusion} by giving conclusive statements and some perspective works.

\section{Related work}
\label{Sec:Related work}

\subsection{Cyberbullying and machine/deep learning}
Many researchers investigated the problem of automatic detection of hate speech in social media networks where various methods, ranging from traditional machine learning approaches to deep learning based approaches, have been put forward \cite{cao2020deephate}. This aims to decrease the risk and consequences of online spread of such harmful content. Besides, many social media platforms such as Facebook and Twitter have made substantial efforts to detect hate speech and stop it from spreading both through manual, semi-manual and fully automated approaches, although with moderate success \cite{cao2020deephate}. Indeed, manual approaches rely on either users' reports of hate speech content, or through hiring individuals whose role is to constantly read the content of the posts and delete or suspend accounts whenever hate speech content is encountered. However, this task remains labor-intensive, time-consuming and inefficient \cite{gamback2017using}, especially when non English content is used, where there is a lack of resources to recognize such toxic content. \citeauthor{poletto2020resources} \cite{poletto2020resources} presented a systematic review of resources and approaches related to hate speech detection in a way to identify gaps in the current research and highlight areas for future works. Toxic content against vulnerable minority groups is investigated by \cite{mossie2020vulnerable} through a combination of word-embeddings and deep-learning algorithms. Automatic detection of cyberbullying incidents has also attracted many interests. For instance, \citeauthor{murnion2018machine} \cite{murnion2018machine} devised a method to automatically collect relevant information from in-game chat data to build a new database from chat messages, while a sentiment analysis technique was then examined as a possible tool for automatic detection of cyberbullying chat messages.
In general, research in cyberbullying discipline tends to focus on messaging, email \cite{chakraborty2018understanding,peled2019cyberbullying}, chat rooms \cite{chandana2018cognitive,chun2020international}, online forums and social media \cite{van2018automatic,yao2019cyberbullying,byrne2018cyberbullying}. Traditional cyberbullying detection is mainly based on supervised algorithms such as Naïve Bayes, SVM, Random Forest, and decision trees \cite{chatzakou2017mean,ducharme2017svm,prathyusha2017cyberbully}. For instance, \citeauthor{balci2015automatic} \cite{balci2015automatic} have examined automatic aggressive behavior detection in online games using machine learning techniques.  
Other approaches have explored combining machine learning and rule-based approaches as in \cite{Foong2017eisic}, or linguistics cues, e.g., negation scope as in \cite{Abderrouaf2019BigData}. 
On the other hand, the success of machine learning based approaches depends on the availability of good quality labeled datasets to build systems capable of properly classifying unseen data, an issue which is often not available. Besides, finding suitable feature engineering is often a challenge by itself, especially given the sparsity of the hate speech corpus.   
Recently, deep learning-based models for cyberbullying detection have overcome many limitations of the traditional machine learning techniques, especially the feature engineering issue, while still providing promising results. For example, \citeauthor{iwendi2020cyberbullying} \cite{iwendi2020cyberbullying} conducted an empirical analysis by comparing four deep learning models; namely, Bidirectional Long Short-Term Memory (BLSTM), Gated Recurrent Units (GRU), Long Short-Term Memory (LSTM), and Recurrent Neural Networks (RNN). This allowed them to determine the effectiveness and performance of deep learning algorithms in detecting insults in Social Commentary. However, deep learning-based approaches require large amounts of good quality data to yield high performances. Data augmentation is one technique to achieve this goal and to enhance the size and quality of training data, which would make the model generalize better. 

On the other hand, it is worth noticing the HatEval task which has been one of the most popular tasks in the SemEval-2019. It aims to gather researchers working towards hate speech detection. Many traditional machine learning and deep-learning based techniques have been introduced such as \cite{baruah2019abaruah,indurthi2019fermi,ameer2019cic,rozental2019amobee}. The participation of Fermi team \cite{indurthi2019fermi} which ranked first relied on the evaluation of multiple sentence embeddings combined to multiple machine learning training models. Pretrained Universal Encoder sentence embeddings were used for transforming the input and SVM is used for the classification.

\subsection{Textual data augmentation}
Different methods could be applied in natural language processing, such as back translation, straightforward data augmentation based on synonym replacement or word embeddings, and NLP augmentation. However, data augmentation in text based tasks and NLP is still challenging. This is due to the nature of text where the basic unit (usually, a word) has both a syntactic and a semantic meaning that depend on the context of the sentence. In general, changing the word may lead to a different meaning and a noisy data which affects the performance of the augmentation technique.For instance, \citeauthor{guzman2020transformers} \cite{guzman2020transformers} compared three data augmentation strategies to deal with aggressiveness detection in Mexican Spanish speech. Namely, Easy Data Augmentation (EDA) \cite{wei2019eda}, Unsupervised Data Augmentation (UDA) \cite{xie2019unsupervised} and Adversarial Data Augmentation (ADA) \cite{jin2020bert}. Similarly, \cite{quteineh2020textual} suggested a new data augmentation approach where artificial text samples are generated through a guided searching procedure and are labeled manually before tackling the classification task. In the same spirit, \citeauthor{rizos2019augment} \cite{rizos2019augment} introduced three data augmentation techniques for text and investigated their performance on three databases using four top performing deep models for hate speech detection. The first technique is based on synonym replacement and word embedding vector closeness. The second one employs a warping of the word tokens along the padded sequence, and the last one deals with a class-conditional, recurrent neural language generation.

On the other hand, paraphrasing is the process of rephrasing the text while keeping the same semantics. Many techniques were used to generate paraphrases in the literature by keeping two main factors: semantic similarity and diversity of generated text \cite{kumar2019submodular}. For instance, \cite{li2017paraphrase,gupta2017deep} focused on obtaining semantically similar paraphrases while \cite{song2018towards,vijayakumar2018diverse} addressed the problem of generating diverse paraphrases. Paraphrasing using back translation seems to be very promising because back-translation models were able to generate several diverse paraphrases while preserving the semantics of the text. 

\begin{algorithm}[htb]
\algsetup{linenosize=\small}
\SetAlgoLined
\SetKwFunction{originDataset}
ltranLabels= tranDataset = []; \\ paraLabels = paraDataset = [];\\
    Read dataset line by line;\\
    Create back translation model;\\
    Create paraphrasing model;\\
    
    \For{line in originDataset}{
    \# get sentence and its label\\
    Label = line('label')\;
    Sentence = line('comment')\;
    
    \# preprocess text and generate new text with back translation\\
    Preprocess(Sentence)\;
    translation = back translation(preprocessed sentence)\;
    Preprocess(translation)\; 
    
    \# expand dataset with back translation\\ 
    tranDataset.append(translation)\;
    tranLabels.append(Label) \;
    
    \# preprocess text and generate new text with paraphrasing\\
    Preprocess(Sentence)\;
    paraphrases = paraphrasing(preprocessed sentence)\;
    Preprocess(paraphrases)\; 
    
    \# expand dataset with paraphrasing\\ 
    \For{paraphrase in paraphrases}{
        paraDataset.append(paraphrases)\;
        paraLabels.append(Label) \;
    }

    \# expand dataset with combination of paraphrasing and back translation\\ 
    
    combinedDataset = tranDataset + paraDataset\;
    combinedLabels = tranLabels + paraLabels\;}
    
    Prepare training and validation sets from multiple configurations:\;
    1- concat(TransDataset, originDataset);\\ 
    2- concat(paraDataset, originDataset);\\ 
    3- concat(combinedDataset, originDataset);\\ 
    Use FastText to create embedding matrix\;
    Create CNN/LSTM network\;
    Classify with CNN/LSTM\;
 \caption{Data Augmentation and hate speech detection : back translation + paraphrasing + CNN}
 \label{Algo:general}
\end{algorithm}

\section{Proposed methodology}
\label{Sec:Methodology}
This paper advocates a combination of back translation and paraphrasing techniques for data augmentation procedure. The augmented data is used later to detect hate speech and cyberbullying contents from social media networks. For this purpose, in a first attempt, we generate synthetic data using back translation. The new data is concatenated to the original data to create a larger dataset. Results at this level are investigated to evaluate the performance of the back translation in generating semantically similar text. In addition, the original data and the newly back translated data are fed to a paraphrasing model. The outcomes of the paraphrase model are also compared to results obtained from the back translated module only. Once the datasets are expanded, preprocessing tools are employed to clean and normalize the text. FastText embeddings are calculated and used as feature set. Finally, a CNN/LSTM is trained on the expanded datasets to classify hate speech/ cyberbullying contents. In summary, our methodology is composed of four steps that we explain in detail in the following subsections and are summarized in Algorithm \ref{Algo:general}. 

\begin{figure}
\centering
\includegraphics[width=0.7\linewidth,height = 8.5cm]{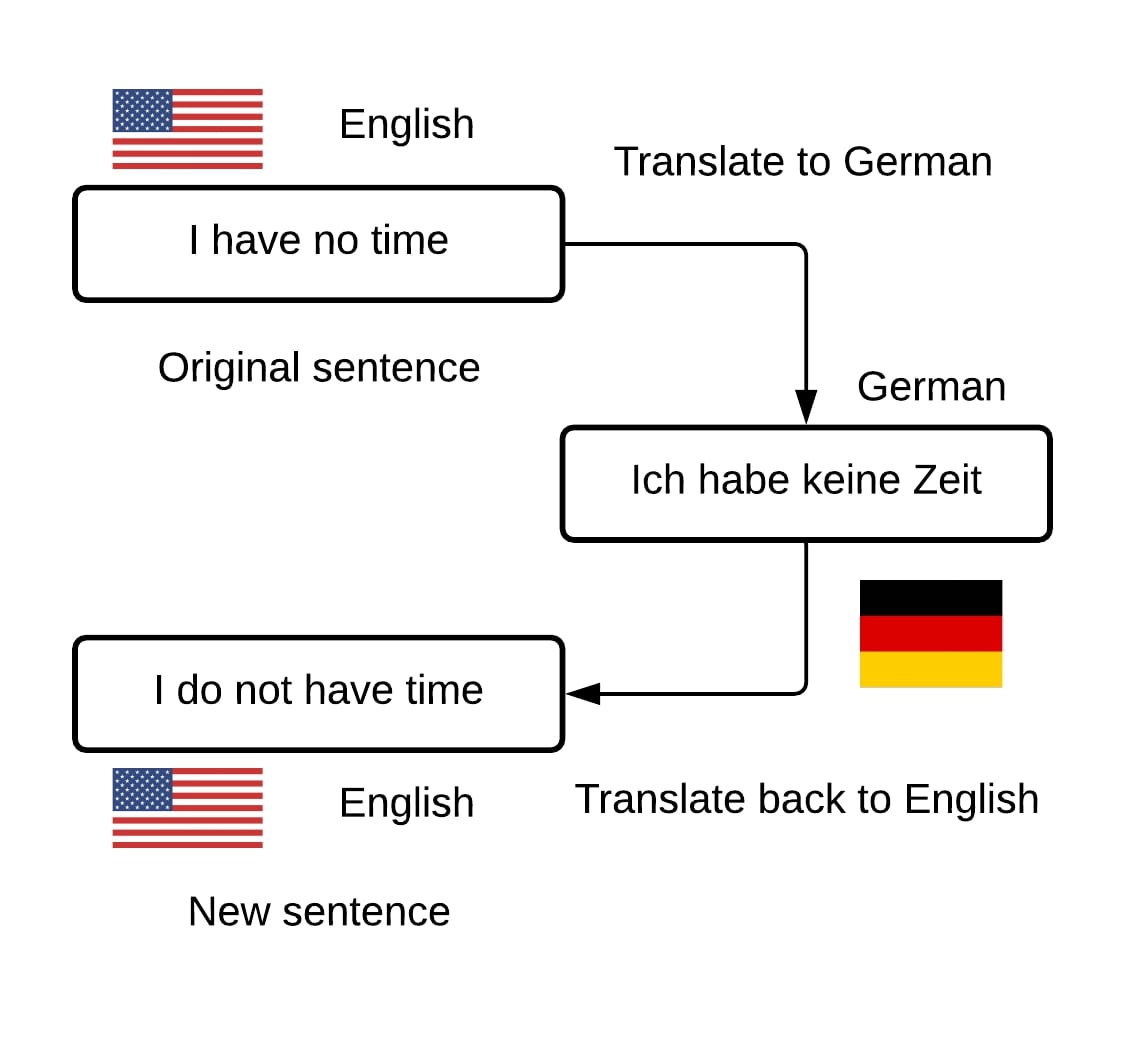}
\caption{Example of generating a new sentence that has the same meaning of the original sentence using back translation.} 
\label{Fig:BackTranslation}
\end{figure}

\subsection{Back Translation module}
\begin{figure}
\includegraphics[width=\textwidth,]{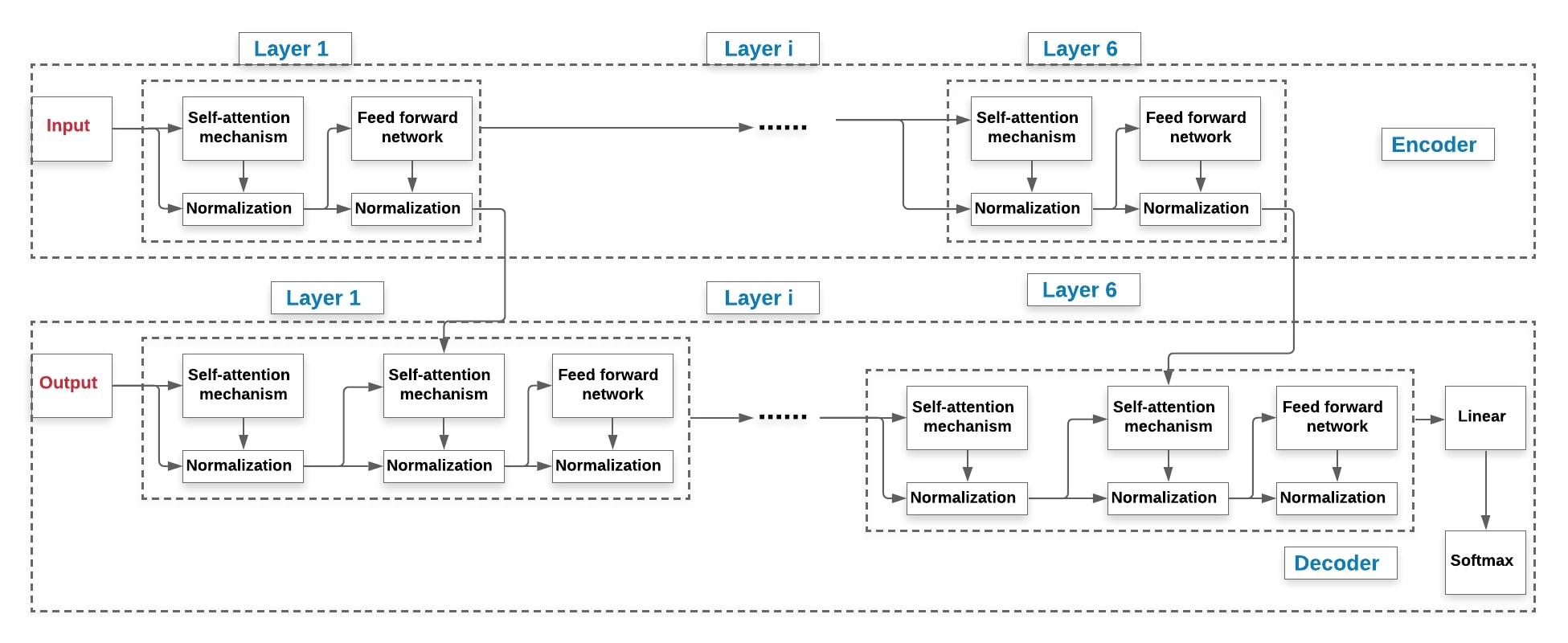}
\caption{The structure of the transformer model used for machine translation tasks. We exploit this architecture for the back translation by duplicating 6 blocks (having same layers) for both the encoder and the decoder.} 
\label{Fig:transformer}
\end{figure}

Back translation augmentation relies on translating text data to another language and then translating it back to the original language. This technique allows generating textual data of distinct wording to original text while preserving the original context and meaning. Fig.~\ref{Fig:BackTranslation} illustrates an example of using back-translation to the sentence: 'I have no time'. Translating this sentence into German and translating it back to English allow us to generate a new sentence. The two sentences still have the same meaning as illustrated in Fig.\ref{Fig:BackTranslation}. 
We exploit the back translation technique provided in \cite{edunov2018understanding,ma2019nlpaug} to generate new sentences close in meaning to our original sentences. This technique was used for neural machine translation at a large scale and yielded promising results. It is based on a pre-trained transformer model (as illustrated in Fig.~\ref{Fig:transformer}), which builds on the architecture presented in \cite{vaswani2017attention}, where six blocks for both encoder and decoder were used. Back-translated sentences based on sampling and noised beam search were generated by translating original data into German and back into English. The model was trained on data from the WMT’18 English-German news translation task \cite{backTranslationWMT18} where the Moses tokenizer was used in line with a joint source and target Byte-Pair-Encoding \cite{sennrich2015improving}. To measure the closeness between the reference human translation and the machine translation, BLEU metric was used. Synthetic data is then concatenated to original sentences to create larger datasets. We provide in Table.~\ref{Tab:samplesBT} some samples of the original sentences and their back translated sentences using our approach. 

\begin{table}
\centering
\caption{Example of back-translated sentences from the used datasets, generated from source sentences with our proposed technique.}
\label{Tab:samplesBT}
\begin{tabular}{|p{80mm}|p{80mm}|}
\hline
\textbf{Source sentence} & \textbf{Back translation}\\
\hline
does it make you mad you have tonsss of hate groups on myspace & does it make you angry that you have countless hate groups on myspace\\
\hline
andrew once again stop talking to my friends nobody likes you you think your hot shit and thats whats annoying bye failed rapist & andrew once again stop talking to my friends no one likes deny you think your horny shit and thats annoying bye failed rapist\\
 \hline
dude are you serious your so stupidd stfu and btw if your tired of people talking shit on your form spring delete this accountt and be yourself are you that bad & dudes are you serious your so stupid stfu and btw if your tired people talk shit on your form spring delete this account and are you yourself so bad\\
  \hline
USER Someone should\'veTaken" this piece of shit to a volcano. & USER Someone should "transport" this piece of shit to a volcano.\\

\hline
\end{tabular}
\end{table}

\begin{table}
\centering
\caption{Example of paraphrases generated from a source sentence after translating it to French and back to English using mixture of experts.}
\label{Tab:Paraphrasing}
\begin{tabular}{|l|p{136mm}|}
\hline
\textbf{Source sentence} & 'Seasons Greetings: Jenna, my very best wishes for the festive season, stay safe and talk to you in 2009.'\\
\hline
\multirow{10}{*}{\textbf{Paraphrases}} & 'Holiday greetings: Jenna, best wishes for the season, stay safe and talk-you in 2009.'\\
    & 'Holiday Greetings: Jenna, my best wishes for the holiday season, please stay safe and speak-you in 2009.'\\
    & 'Holiday Greetings: Jenna, my best wishes for the holiday season, stay safe and talk-you in 2009.'\\
    & 'Holiday greetings: Jenna, holiday season best wishes, stay safe and talk-you in 2009.'\\
    & 'Holiday greetings: Jenna, best wishes for the holiday season, stay safe and keep talking-you in 2009.'\\
    & 'Holiday greetings: Jenna, my best wishes for the season, stay safe and speak-you in 2009.'\\
    & 'Holiday greetings: Jenna, my best wishes for the holiday season, stay safe and talk-you in 2009.'\\
    & 'Holiday greetings: Jenna, my best wishes for the holiday season, stay safe and stay-you in 2009.'\\
    & 'Holiday greetings: Jenna, my best wishes for the season, stay safe and speak-you in 2009.'\\
    & 'Greetings from the holidays: Jenna, my best wishes for the season, stay safe and speak-you in 2009.'\\
\hline
\end{tabular}
\end{table}

\subsection{Paraphrasing module}
As mentioned earlier in this paper, paraphrasing using back translation can yield useful equivalent representations of the original text as well. It aims at rewriting the original text in different words without changing its meaning by passing through an intermediate language. This can be used for many purposes such as text simplification, document summarization, data augmentation and information retrieval. We exploit machine translation models \cite{ott2019fairseq} to translate English text into French and back to English (round-trip translation). The French-English translation model is based on a mixture of experts as provided by \cite{shen2019mixture}. Examples of paraphrases generated from an input sentence are illustrated in Table.\ref{Tab:Paraphrasing}.

The generated paraphrases are then concatenated to the original sentences to create an initial expanded dataset. Finally, the data constructed from the back translation augmentation and the data generated from the paraphrasing augmentation are concatenated to obtain a final expanded dataset.   

\subsection{Preprocessing and Feature Extraction}
To improve the performance of our proposal while encoding the data, it is important to perform some preprocessing steps to clean out the text and make it more useful for subsequent analysis. For that, we started by removing stop words, uncommon characters, symbols, and punctuation. Then, we replaced abbreviations with their original wording. Afterwards, we transformed text into numeric form to make it more digestible by the machine learning algorithms. So, we calculated word-embeddings that represent our sentences using sequences of real number vectors instead of word sequences using the pre-trained word embeddings FastText \footnote{\footnotesize\url{https://s3-us-west-1.amazonaws.com/fasttext-vectors/wiki-news-300d-1M.vec.zip} (accessed March, 2021)}. 

We choose FastText embeddings because it enables us to improve the vector representations obtained with the skip-gram method by using internal structure of the words. Unlike Word2Vec and Glove, FastText can generate embedding for words that do not occur in the training corpus by relying on the character representation only, which enables us to bypass out-of-word dictionary.  Therefore, FastText embeddings constitute our features for hate speech and cyberbullying detection. 

\subsection{Classification Architecture}
\label{Sec:Classification Setup}
\begin{figure}
\includegraphics[width=\textwidth]{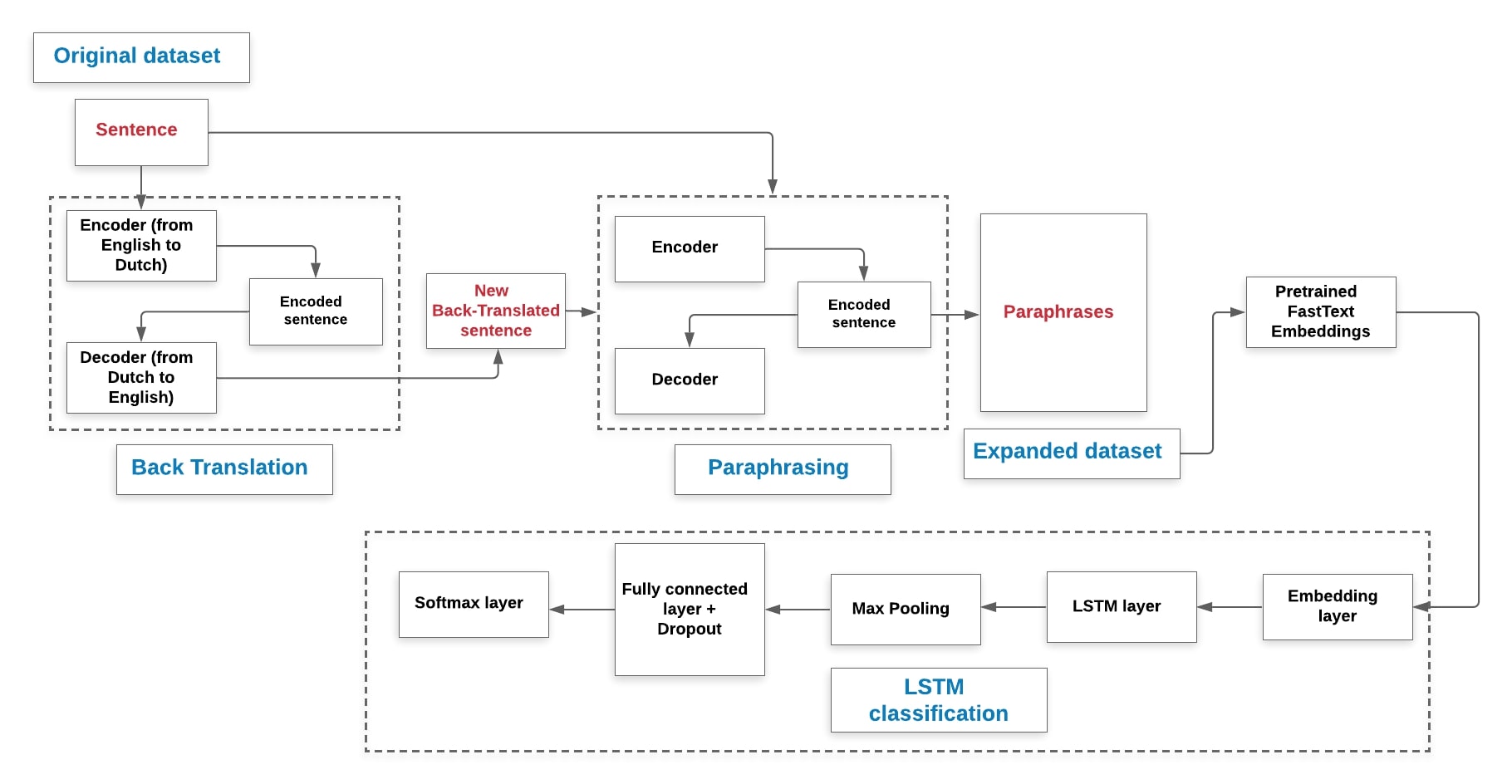}
\caption{General scheme of our proposal where FastText embeddings from expanded dataset generated using back translation and paraphrasing are calculated. Then, hate speech is classified using an LSTM model. In this figure we illustrate the architecture of the LSTM network.} 
\label{Fig:lstm}
\end{figure}

Once our feature vectors are calculated, we perform the classification task either using the CNN architecture. Initially, we employed a random split of the original dataset into 70\% for training and 30\% for testing and validation. To provide a balanced training, the same proportion of dataset has been used for all kinds of datasets (both original and artificially generated datasets). We employed Kim Convolution Neural Network (CNN) architecture \cite{kim2014convolutional} as illustrated in Fig.\ref{Fig:general}, where the input layer is represented as a concatenation of the words forming the post (up to 70 words), except that each word is now represented by its FastText embedding representation with a 300 embedding vector. A convolution 1D operation with a kernel size of 3 was used with a max-over-time pooling operation over the feature map with a layer dense 50. Dropout on the penultimate layer with a constraint on l2-norm of the weight vector was used for regularization. 
Furthermore, we also contrasted the CNN architecture with LSTM model for comparison purpose. 
For the LSTM model we followed \citeauthor{van2018challenges} (as shown in Fig.\ref{Fig:lstm}), where an 128 unit LSTM layer is used inline with the embedding layer and dropout. Finally, a fully connected layer with softmax is employed to leverage the class of each entry into hate speech or not hate speech. The input of our LSTM model is the pretrained Fastext embeddings.

Furthermore, we have used two baseline models: Linear regression (LR) and Naive Bayes (NB) to compare our experiment's results. As features, we have used TF-IDF word-level, n-gram word level, and n-gram character level for NB and LR with n=2. Since the n-gram character level performed better than others, in the result section, we have shown only the n-gram character level feature with n=2 for the baseline model as we found it to achieve better performances.

\section{Experimental results}
\label{Sec:Results}

We evaluate our approach of cyberbullying detection on five publicly available datasets: AskFm, Formspring, OLID 2019, Warner and Waseem, and Wikipedia toxic comments dataset. Table.\ref{Tab:datasets} summarizes some important information about the used datasets including the size, number of classes, number of samples per class and some samples from each dataset. In addition, in the subsequent sections, we present a brief description of these datasets, followed by the performance evaluation metrics and our experimental results.

We compare the results of each original dataset to the expanded one. In addition, results calculated in terms of accuracy, F1 score, recall and precision have been reported and summarized in Tables~\ref{Tab:results}, \ref{Tab:results2}, and \ref{Tab:recallprecision}. 

\begin{table}
\caption{Some important information on the used datasets with some samples from the hate speech class}
\centering
\label{Tab:datasets}
\begin{tabular}{|p{20mm}|l|l|p{25mm}|p{75mm}|}
\hline
Dataset & Size & Classes & Number of samples per class & Examples\\
\hline
\textbf{AskFm} & 9998 & 2 classes & Cyberbullying: 1209 

Not cyberbullying: 8789  & *'pointing out that muslims are useless parasites on the west is also not going to make them change their ways because that s what they are unashamedly here for'

*'no india can code an entire software and whatever he codes is buggy and if he does the fonrt end that will be like 4th world shit that is india hahaha'

*'why the sand niggers should vote for obama'\\
\hline
\textbf{Formspring}  & 12772 & 2 classes & Hate: 892 

Not hate: 11880 &  *'why do you follow the bandwagon Formspring twitter whats next gay justin beiber' 

*'dude are you serious your so stupidd stfu and btw if your tired of people talking shit on your form spring delete this accountt and be yourself are you that bad'\\
\hline
\textbf{Olid}  & 14100 & 3 levels of classes & Offensive: 4384

Not offensive: 9716 & *'user i mean it worked for gun control right url'

*'user user oh noes tough shit'

*'user user obama wanted liberals amp illegals to move into red states'\\
\hline
\textbf{Warner and Waseem}  & 1870 & 2 classes & Hate: 257 

Not hate: 1613 & *'pointing out that muslims are useless parasites on the west is also not going to make them change their ways because that s what they are unashamedly here for'

*'official nigger owner s manual'\\
\hline
\textbf{Toxic Wikipedia} & 159571 & 7 classes & Toxic: 976

Not toxic: 8024& *'to donate to this group of buffoons'

*'i hate you i hate you'

*'dude this guy is so gay'\\
\hline
\end{tabular}
\end{table}

\subsection{Datasets} 
\paragraph{AskFm dataset}: contains data from Ask.fm website \footnote{\footnotesize\url{https://ask.fm/}} which serves for asking questions either anonymously or non-anonymously, and, then, retrieving answers from users. This dataset was employed to detect cyberbullying in question-answer pairs and is composed of 10000 entries, where label 1 is assigned if the entry corresponds to cyberbullying and 0 otherwise. 

\paragraph{Formspring dataset}: contains 12.772 samples of data from the question-and-answer-based social network Formspring.me \cite{reynolds2011using}, which is available on Kaggle. Unlabeled data was collected in the summer of 2010 where only 7\% was labeled as cyberbullying. The Formspring service was shutdown in 2013 due to cyberbullying-related deaths of few teenagers in 2011. See, also \cite{Jahan2020Semeval} for a detailed analysis of AskFm and Formspring dataset.   

\paragraph{Olid 2019 dataset} \cite{zampierietal2019}: contains a collection of crowd-sourcing annotated tweets that can be used for offensive language detection. The dataset contains 14100 samples split into a 13240 tweets for training for which 3\% correspond to offensive language, and a testing set of 860 tweets. It has been used in the OffensEval-2019 competition in SemEval-2019\cite{zampieri2019semeval}. It is a hierarchical dataset for which three levels are considered. First, one distinguishes Offensive / Not-Offensive level. Second, if offensive, then a tweet is categorized whether it is a Targeted-Insult or Untargeted-Insult. Third, the target is further discriminated depending whether it stands for an individual, a group, or some other objects.

\paragraph{Warner and Waseem dataset}: contains approximately 2000 samples of Twitter data. The dataset is a sub-dataset of "Warner" (\citeauthor{warner2012detecting}, 2012) and "Waseem" (\citeauthor{waseem2016hateful}, 2016), which is publicly available in  \footnote{\url{https://github.com/uds-lsv/lexicon-of-abusive-words} (accessed April 20, 2021)} and has only abusive micro-posts that were considered to be explicitly abusive (i.e., there is at least one abusive term included). Since this dataset considers at least one abusive term, this makes it different from other datasets.  

\paragraph{Wikipedia toxic classification dataset}: was created for the Toxic Comment Classification Challenge 2017 on Kaggle. It has been released by the Conversation AI team, a research initiative founded by Jigsaw and Google \footnote{\footnotesize\url{https://www.kaggle.com/c/jigsaw-toxic-comment-classification-challenge} (accessed May 15, 2021)}. It contains 159,571 comments from Wikipedia where different labels were used to classify each data entry into multiple categories: toxic, severe\_toxic, obscene, threat, insult, and identity\_hate. We are interested to only one category (toxic), so we omit the other labels. This dataset contains many challenges related to NLP and could be used to detect toxic speech. In addition, it features an unbalanced class distribution. We use only 9000 entries from this dataset.

\subsection{Performance metrics} 
To demonstrate the performance of our proposal, we calculate the accuracy and F\_Measure as follows: 

\textit{F\_Measure}: determines the harmonic mean of precision and recall by giving information about the test’s accuracy. It is expressed mathematically as:

\begin{equation} 
F\_Measure   = 2* \frac{Precision * Recall}{Precision + Recall}
\end{equation}

\textit{Accuracy}: measures the percentage of correct predictions relative to the total number of samples. It can be expressed as:

\begin{equation}  
Accuracy   =  \frac{\text{Correct Predictions}}{\text{Total Predictions}} = \frac{TP + FN} {TP + FN + TN + FP}
\end{equation} 

\textit{Recall}: measures the proportion of data predicted in its class. Mathematically, this can be expressed as:

\begin{equation}  
Recall   =  \frac{TP } {TP + FN}
\end{equation} 

\textit{Precision}: measures the likelihood of a detected instance of data to its real occurrence. It can be calculated as follows:

\begin{equation}  
Precision   =  \frac{TP } {TP + FP}
\end{equation} 

Where TP, FN, TN, FP correspond to true positive, false negative, true negative and false positive respectively. 

\begin{table}[htb]
\centering
\caption{Size of datasets before and after expansion using different configurations of back translation, paraphrasing using a mixture of experts translation models and combination of both augmentation techniques. BT refers to back translation.}\label{tab2}
\begin{tabular}{|l|l|l|l|l|l|}
\hline
Dataset & AskFM & Formspring & Olid & Warner and Waseem  & Wikipedia Toxic\\
\hline
\textbf{Before expansion}& 10K & 12K & 13K & 1.8K & 9K\\
\textbf{After BT} & 20K & 25.4K & 26.5K & 3.7K &  18K\\
\textbf{After paraphrasing}& 94.3K & 130.5K & 132.6K & 20K & 89K \\
\textbf{BT + Paraphrasing} & 188.7K & 261K & 265.2K & 41K & 172K \\
\hline
\end{tabular}
\end{table}

\subsection{Results} 
We first investigate the size of the generated datasets using different configurations before examining the results of hate speech detection. We show in Table~\ref{tab2} the size of the original and the augmented datasets using back translation, paraphrasing and combination of both of them. As illustrated in this table, original datasets have been expanded up to twice the original dataset when using the back translation technique. Similarly, the size of the augmented dataset is around $\times$ 10 of the original dataset size while using the paraphrasing on all datasets. Again, the sizes of the augmented datasets after combining the back translation and the paraphrasing methods are approximately $\times$ 20 of the original datasets, $\times$ 10 of the augmented datasets with back translation and $\times$ 2 of the augmented datasets with paraphrasing only.

Once the augmentation techniques are investigated in terms of datasets sizes, we evaluate the performance of using these augmented datasets compared to original ones for detecting hate speech and cyberbullying. From Table~\ref{Tab:results}, we can see that the classifier accuracy and F1 score have increased by 1-6 percent (\%) for the expanded five datasets using back translation. Among all three classifiers, CNN performs best for all initial datasets and their extended datasets. CNN performance improved 2-6\% after the data augmentation. In contrast, the baseline classifiers improved by 0-3\%. This indicates that the adopted CNN architecture with the back translation augmentation strategy worked better than the baseline classifiers. However, as we can see, in all cases, the proposed method worked for both the neural network and the baseline models, which indicates the feasibility of the use of the proposed methods.

\begin{table}
\caption{Performance metrics in terms of Accuracy (\%) and F1 scores (\%) of CNN with FastText, LR and NB with n-gram (n=2) character level classification representation for original and expanded datasets.}
\label{Tab:results}
\centering
\begin{tabular}{|l|l|l|l|l|l|l|l|}
\hline
\bf{} & \bf{} & \multicolumn{2}{c|}{\bf{CNN}} & \multicolumn{2}{c|}{\bf{LR}}&\multicolumn{2}{c|}{\bf{NB}} \\
\bf{Dataset Name}  & \bf{Experiment with}  & \bf{Acc.}  & \bf{F1} & \bf{Acc.}  & \bf{F1}& \bf{Acc.}  & \bf{F1} \\ \hline 

\multirow{2}{12em} {\textbf{AskFm dataset}} & Not expanded dataset & 90.3 & 88.9 & 91.2 & 89.3 & 88.2 & 83.0 \\ 
& Expanded dataset & 94.8 & 94.8 & 92.1 & 90.2 & 87.8& 85.3\\\hline 

\multirow {2}{12em}{\textbf{FormSpring dataset}} & Not expanded dataset & 95.2 & 94.7 & 94.9 & 93.7 & 93.5& 90.8\\ 
& Expanded dataset & 98.2 & 98.1 & 96.3& 95.9& 95.8& 93.2 \\\hline 

\multirow {2}{12em}{\textbf{OLID 2109 dataset}} & Not expanded dataset & 78.1& 77.4& 75.2& 72.8& 71.3& 64.3 \\ 
& Expanded dataset & 83.1& 82.1 & 77.3& 75.1 & 72.4& 67.3 \\\hline  

\multirow {2}{12em} {\textbf{Warner and Waseem Dataset}} & Not expanded dataset & 90.2& 89.4& 89.3& 85.1& 88.4& 83.1 \\ 
& Expanded dataset & 92.0 & 92.3 & 90.4 & 87.4& 88.2 & 83.3\\\hline  

\multirow {2}{12em} {\textbf{Wikepedia Dataset}} & Not expanded dataset & 95.5 & 95.3 & 93.7& 92.4 & 92.3 & 90.0  \\ 
& Expanded dataset &  99.4 & 99.4 & 95.5 & 95.1 & 92.9 & 91.8    \\\hline

\end{tabular}
\end{table}

\begin{table}[]
\caption{Performance metrics in terms of Accuracy (\%) and F1 scores (\%) of CNN and LSTM classifications with FastText representation for original and expanded datasets.}
\label{Tab:results2}
\centering
\begin{tabular}{|l|l|l|l|l|l|l|}
\hline
\bf{} & \bf{} & \multicolumn{2}{c|}{\bf{CNN}}  & \multicolumn{2}{c|}{\bf{LSTM}}\\
\bf{Dataset Name}  & \bf{Experiment with}  & \bf{Acc.}  & \bf{F1} & \bf{Acc.}  & \bf{F1}\\ \hline 

\multirow {2}{12em} {\textbf{AskFM Dataset}} & Not expanded dataset & 93.7 & 93.4 & 92.1  & 91.3\\ 
& Expanded dataset with back translation & 96.9 & 96.9 & 93.9 & 93.9\\
& Expanded dataset with paraphrasing & 98.7 & 98.7 & 100 & 100\\
& Expanded dataset with back translation + paraphrasing & 98.9 & 98.8 & 98.4 & 98.4\\\hline  

\multirow {2}{12em} {\textbf{Formspring Dataset}} & Not expanded dataset & 96.6 & 96.4 & 96.2 & 95.9\\ 
& Expanded dataset with back translation & 98.4 & 98.4 & 97.0 & 96.7\\
& Expanded dataset with paraphrasing & 99.2 & 99.2 & 98.7 & 98.7 \\
& Expanded dataset with back translation + paraphrasing & 99.4 & 99.4 & 99.2 & 99.1\\\hline  

\multirow {2}{12em} {\textbf{Olid Dataset}} & Not expanded dataset & 71.6 & 71.0 & 74.5 & 73.5\\ 
& Expanded dataset with back translation & 84.7 & 84.5  &  78.0 & 77.4\\
& Expanded dataset with paraphrasing & 96.0 &  96.0 & 95.2 & 95.1\\
& Expanded dataset with back translation + paraphrasing & 94.4 & 94.5 & 96.4 & 96.4\\\hline  

\multirow {2}{12em} {\textbf{Warner and Waseem Dataset}} & Not expanded dataset & 90.0 & 89.0 & 90.1 & 89.1\\ 
& Expanded dataset with back translation & 92.3 & 92.0 & 92.8 &92.1\\
& Expanded dataset with paraphrasing & 98.5 & 98.5 & 99.1 & 99.1\\
& Expanded dataset with back translation + paraphrasing & 99.3 & 99.3 & 99.2 & 99.2\\
\hline 

\multirow {2}{12em} {\textbf{Wikipedia Toxic comments Dataset}} & Not expanded dataset & 95.5 & 95.3 & 95.6 & 95.5\\ 
& Expanded dataset with back translation & 97.3 & 97.3 & 95.9 & 95.7\\
& Expanded dataset with paraphrasing & 99.1 & 99.1 & 99.2  & 99.2\\
& Expanded dataset with back translation + paraphrasing & 99.4 & 99.4 & 98.1 & 98.0\\\hline    
\end{tabular}
\end{table}

\begin{table}
\caption{Performance metrics in terms of Recall (\%) and Precision (\%) of CNN with FastText classification representation for original and expanded datasets.}
\label{Tab:recallprecision}
\centering
\begin{tabular}{|l|l|l|l|l|l|}
\hline
\bf{} & \bf{} & \multicolumn{2}{c|}{\bf{CNN}} & \multicolumn{2}{c|}{\bf{LSTM}}  \\
\bf{Dataset Name} & \bf{Experiment with} & \bf{Recall} & \bf{Precision} & \bf{Recall} & \bf{Precision} \\
 \hline 

\multirow{2}{12em} {\textbf{AskFm dataset}} & Not expanded dataset & 93.6 & 93.2 & 91.9 & 91.2\\ 
& Expanded dataset & 98.8 & 98.8 & 98.6 & 98.6\\\hline 

\multirow {2}{12em}{\textbf{FormSpring dataset}} & Not expanded dataset & 96.6 & 96.6 & 96.4 & 96.1\\ 
& Expanded dataset & 99.3 & 99.2  & 99.2 & 99.1\\\hline 

\multirow {2}{12em}{\textbf{OLID 2109 dataset}} & Not expanded dataset & 70.7 & 70.3 & 74.3 &73.6\\ 
& Expanded dataset & 96.2 & 96.3 & 96.3 & 96.3\\\hline  

\multirow {2}{12em} {\textbf{Warner and Waseem Dataset}} & Not expanded dataset & 92.5 & 92.1 & 89.3 & 87.5\\ 
& Expanded dataset & 99.7 & 99.6 & 99.3 & 99.2\\\hline  

\multirow {2}{12em} {\textbf{Wikepedia Dataset}} & Not expanded dataset & 95.3 & 95.2 & 95.6& 95.4\\ 
& Expanded dataset & 98.0 & 97.9 & 98.1 &98.0\\\hline  

\end{tabular}
\end{table}

\begin{table}[htb]
\centering
\caption{Results of CNN classification while augmenting only the minority class using the back translation compared to results of augmenting the whole dataset in terms of Accuracy (\%) }\label{Tab:minority}
\begin{tabular}{|l|l|l|}
\hline
Dataset &  Whole dataset & Minority class\\
\hline
\textbf{Ask.fm } & 96.9  & 91.3\\
\textbf{Formspring } & 98.4  & 96.2\\
\textbf{OLID-2019 } & 84.7 & 71.1 \\
\textbf{Warner and Waseem } & 92.3 & 90.8 \\
\textbf{Wikipedia toxic } & 97.3 & 94.5 \\
\hline
\end{tabular}
\end{table}

\begin{table}
\centering
\caption{FormSpring datasets results (accuracy in \%) comparison using CNN acrhitecture between \cite{zhang2016cyberbullying} and ours expanded FormSpring datastes}
\label{Tab:resultsComparison}
\begin{tabular}{|p{2.7cm}|l|l|l|}
\hline
\bf{Authors name} & \bf{Classifier} & \bf{Accuracy} & \bf{F1 score}
\\ \hline

Zhang & CNN  &  96.4  & 48.0  \\
\hline
Zhang & PCNN  & 96.8   & 56.0  \\
\hline
Ours (Expanded FormSpring dataset) & CNN   & 98.0 & 98.0 \\\hline
\end{tabular}
\end{table}

Similarly, from Table~\ref{Tab:results2}, we can see that the classifier accuracy and F1 score have increased by 4-10 percent (\%) for the four expanded datasets using paraphrasing and the combination of back translation and paraphrasing with the CNN classifier. The results obtained after combining back translation and paraphrasing are better than that using only one of the techniques for augmenting the data, only for Olid dataset for which results of paraphrasing are better than the combination of both techniques. However, the paraphrasing performed much better than the back translation and its results are very close to the results obtained after the combination. This could be justified by the diversity and quantity of generated paraphrases that, as we mentioned before, are close in meaning to the original data. We observed the same thing when classifying data using an LSTM network. The accuracy and F1 score have increased by 3-23 percent (\%). Results of classification of data expanded with combination of back translation and paraphrasing are better than other configurations, only for AskFM dataset for which results of paraphrasing are better than the combination of both techniques. Moreover, the CNN classifier performed better than the LSTM, only for the Olid dataset. However, the results of both classifiers are close and the difference between their performances is negligible.

\begin{figure}
\centering
\subfloat[]{\includegraphics[height=2.35in, width=3.2in]{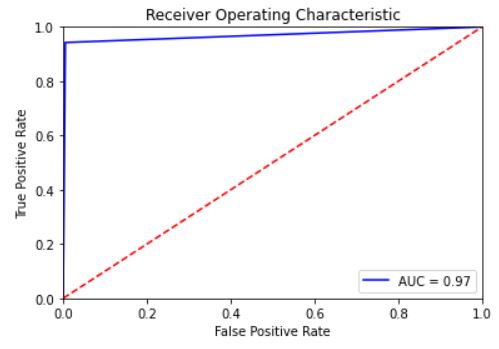}
\label{subfig:ROCAskFM}}
\hfil
\subfloat[]{\includegraphics[height=2.35in,width=3.2in]{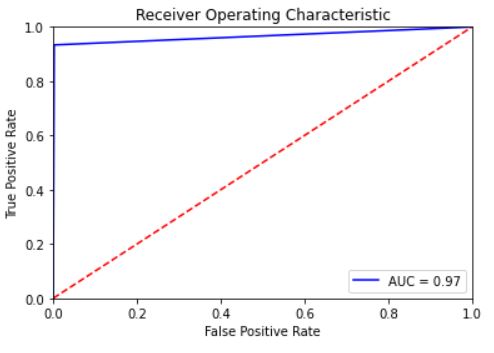}
\label{subfig:ROCFormspring}}

\subfloat[]{\includegraphics[height=2.35in, width=3.2in]{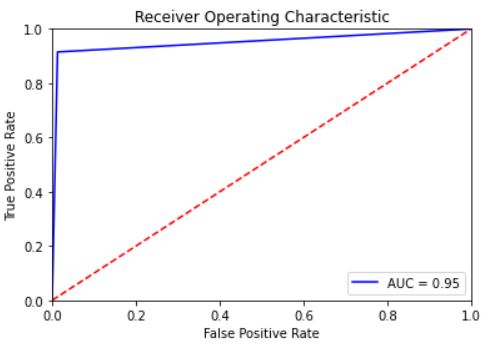}
\label{subfig:ROCOlid}}
\hfil
\subfloat[]{\includegraphics[height=2.35in,width=3.2in]{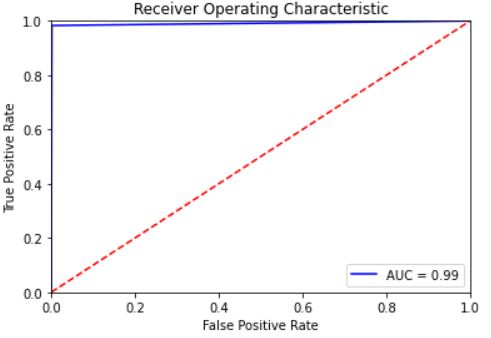}
\label{subfig:ROCWaseem}}

\subfloat[]{\includegraphics[height=2.4in,width=3.2in]{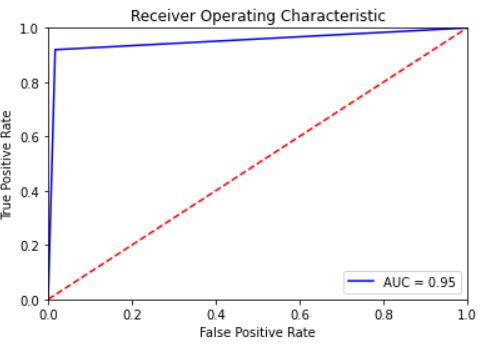}
\label{subfig:ROCWiki}}

\caption{ROC curve for hate speech detection using CNN for the expanded datasets: a. the AskFm dataset, b. the Formspring dataset, c.the Olid dataset, d. the Warner and Waseem dataset, and e. the Wikipedia toxic comment dataset.}
\label{fig:ROC}
\end{figure}

\begin{figure}
\centering
\subfloat[]{\includegraphics[height=2.35in, width=3.2in]{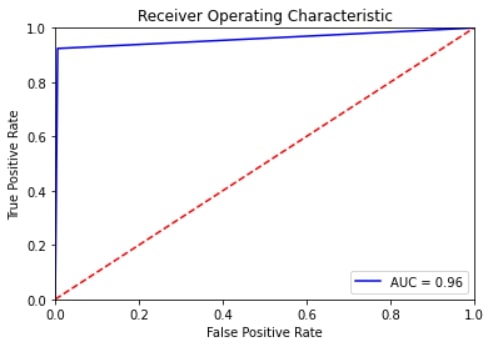}
\label{subfig:ROCAskFM1}}
\hfil
\subfloat[]{\includegraphics[height=2.35in,width=3.2in]{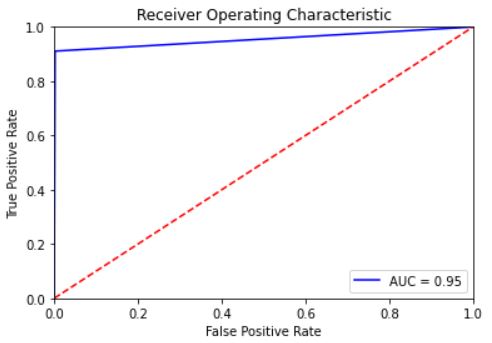}
\label{subfig:ROCFormspring1}}

\subfloat[]{\includegraphics[height=2.35in, width=3.2in]{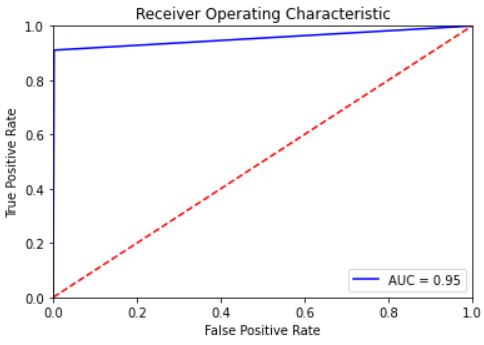}
\label{subfig:ROCOlid1}}
\hfil
\subfloat[]{\includegraphics[height=2.35in,width=3.2in]{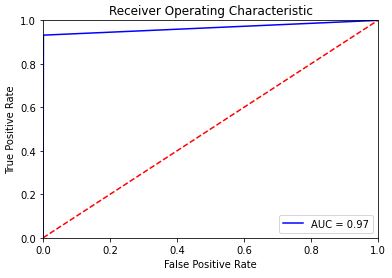}
\label{subfig:ROCWaseem1}}

\subfloat[]{\includegraphics[height=2.4in,width=3.2in]{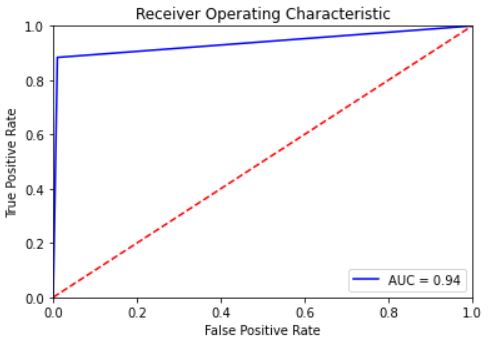}
\label{subfig:ROCWiki1}}

\caption{ROC curve for hate speech detection using LSTM for the expanded datasets: a. the AskFm dataset, b. the Formspring dataset, c.the Olid dataset, d. the Warner and Waseem dataset, and e. the Wikipedia toxic comment dataset.}
\label{fig:ROCLSTM}
\end{figure}  

For the dataset Warner and Waseem, F1 scores have improved 3\% for both CNN and LSTM architectures with back translation and 9\% with paraphrasing, even though the initial dataset was very small, only 1.8k. On the other hand, the other three datasets, AsFfm, FormSpring, and Olid dataset, could be considered much larger than the Warner and Waseem dataset. However, data augmentation showed performance improvement in both accuracy and F1 score. This validates the use of the back translation method for small to a large set of datasets. 

From another perspective, we compared the recall and the precision values for original and expanded datasets using the combination of back-translation and paraphrasing in Table~\ref{Tab:recallprecision}. We observed that for the CNN classifier, the recall and precision increased by 2-26 percent (\%) after applying our data augmentation technique. Similarly, for the LSTM model, the recall and the precision increased by 3-23 percent (\%) from the original to the augmented dataset. Figure~\ref{fig:ROC} and \ref{fig:ROCLSTM} illustrate the ROC curve for hate speech classification using the expanded datasets. Both show the trade-off between sensitivity (or TPR) and specificity (1 – FPR). As we can notice, the curves for all datasets are closer to the top-left corner which indicates the soundness of our CNN and LSTM classifiers in classifying hate speech instances among normal cases. The best ROC curve was obtained for the Warner and Waseem dataset for both classifiers with an AUC value of 99\% for the CNN and 97\% for the LSTM. In addition, AUC values for all datasets with LSTM classifier are marginally smaller than their corresponding values for the CNN classifier. 

Indeed, the LSTM and the CNN models performed well for hate speech classification and yielded good results in terms of accuracy, F1 score, recall and precision. For unbalanced class datasets, recall and precision could provide efficient and accurate metrics about the performance of the classifier.

The first two datasets (AskFm and Formspring) were mainly based on cyberbullying. The other datasets were related to hate speech. However, we have acknowledged a non-negligible performance improvement in both cases, suggesting that our proposed methods could be used for different NLP challenges as well (i.e., word-sense disambiguation, automatic translation, automatic text summarization, among others).

Furthermore, we compared results of hate speech classification by augmenting the whole dataset (both classes: hate and not hate) versus augmenting only the minority class (hate class) with back-translation. We present in Table~\ref{Tab:minority} only the results of the CNN classification since it performed better than the LSTM model. We observe that, for all datasets, augmenting the whole dataset gave much better results than augmenting the hate class where the margin observed is between 2\% and 13\%. The largest margin of 13\% was observed for the Olid dataset and this is due to the percentage of hate class data that is 31\% and is the biggest among other datasets.

Similarly to us, Zhang et al. \cite{zhang2016cyberbullying} proposed a novel cyberbullying detection using a pronunciation based convolutional neural network (PCNN) architecture. Since they used Formspring datasets, a light comparison to our results is reported in Table~\ref{Tab:resultsComparison}. The comparison of these results clearly shows that both CNN model and PCNN model of \cite{zhang2016cyberbullying} achieve a max of 96.8\% accuracy and 56\% F1 scores. However, our CNN model trained on expanded Formspring datasets using back translation  yields 1.8\% higher accuracy and 42\% higher F1 scores.

\subsection{Error analysis}
Although we have achieved quite highly accurate results in terms of accuracy and F1 score, the model still exhibit a negligible portion of false detection. To understand this phenomenon better, we performed in this section a deep analysis of the model's error. 

For this purpose, we randomly prepared 3 subsets of test data from the previous experiments (AskFm, FormSpring, and Warner-Waseem), then manually inspected the classifier output. Each test data contained 100 samples among which 80 were non-hate speech and 20 were hate samples. Here AskFm and FormSpring datasets were related to cyberbullying, but for consistency, we used the terms hate and non-hate for all three datasets. Our experimental setup and sample test data were identical for both non-expanded and expanded datasets. Since our results showed CNN classifier out-performing others, we considered only CNN for error analysis. Table \ref{Error} presents error analysis results for both non-expanded and expanded datasets.

In \textbf{AskFm dataset}, it was evident that none of the non-hate samples was misclassified as a hate; however, only samples belonging to the hate classes were misclassified as non-hate. Before data augmentation, models' success rate was 100\% for non-hate and dropped to 55\% for hate samples. However, after data augmentation, the accuracy of detecting hate samples improved to 85\% and hold its previous 100\% success for non-hate detection. In a total of 100 samples, only 3 samples were misclassified as non-hate while it was initially labeled as hate. Among these 3 samples, some words (i.e., thick, drive, Fuck) can have multiple meanings (sexual or non-sexual sense) based on the context. Besides, the word 'lamo', which is a social slang, is usually used for fun rather than serious conversation. For example, the sample sentence 'lmao when u studpid fuck' was misclassified as a non-hate, and this might look as a straight error. However, if we further examine the issue, we may notice that the post has several elements that might deceive the classifier (e.g., stupid wrongly spelled as 'studpid',  'u' used as short form of 'you', and 'lamo' often used for fun).

\begin{table}[htb]
\centering
\caption{ Error analysis of  CNN classifier before and after dataset augmentation. Accuracy (\%). }
\label{Error}
\begin{tabular}{|p{1.7cm}|p{1.3cm}|p{1.3cm}|p{1cm}|p{10cm}|}
\hline
Dataset & Non-hate detection Acc. & Hate detection Acc.& Overall Acc. & Misclassified sentence\\
\hline

\textbf{AskFm } & 100 &  55 & 89 & 
\textbf{Nine hate samples   misclassified as non-hate, e.g.:}\newline
1. this person is 100 fake so beware isabella \newline
2. lmao he was thick tho \newline
3. no he a bitch \newline
4. eyes but i here you have a nice butt to how about driving from back \newline
5. how short is your shortest shorts i head its small \newline
6. lmao when u studpid fuck
\\
\hline

\textbf{AskFm Expanded BT + paraphrasing} & 100 &  85 & 97 & 
\textbf{Three hate samples   misclassified as non-hate:}\newline
1. lmao he was thick tho  \newline
2. eyes but i here you have a nice butt to how about driving from back \newline
3. lmao when u studpid fuck
\\
\hline

\textbf{FormSpring } & 93 &  75 & 91 & 
\textbf{Five non-hate samples   misclassified as hate, e.g.:}\newline   
1. great time with you gys: \newline
2. go read my Formspringr ahora \newline
\textbf{Four hate samples   misclassified as non-hate, e.g.:}\newline   
1. whyy aree youu suchh aa faggot \newline
2. love you sexy thang  
\\
\hline
\textbf{FormSpring Exp. BT + paraphrasing } & 98.7 &  90 & 97 & 

\textbf{One non-hate sample   misclassified as hate:}\newline
1. Dont do her you bad   \newline
\textbf{Two hate samples   misclassified as non-hate:}\newline
1. love you sexy thang   \newline
2. how old are you and how do you know michael
\\
\hline

\textbf{Warner and Waseem} & 100 &  15 & 83 & 
\textbf{Seventeen hate samples   misclassified as non-hate, e.g.:}\newline
1. what s the difference between a dead dog in the road and a dead nigger\newline
2. kill all niggers oops wrong forum but hey i aint racist kill all foreighners\newline
3. Six feet chink posted june 
\\
\hline

\textbf{Warner and Waseem Exp. BT + Paraphrasing} & 100 &  95 & 99 & 
\textbf{One hate sample   misclassified as non-hate:}\newline
1. dis nigga is real about his an shit. i read ppls post on here an on some real shit. white ppl get mad when we got ta nigga like dis cuz niggaz like dis will tell them that tru shit an not that bullshit an if white ppl get mad over dis o will shit look wat thay did to our ppl back in da day an hold up on that note its not just us. its everybody white ppl think thay can get over on ppl but not nomore cuz thay know if u fuck whit somebody u dont know ur not just fuckin wit one preson but u fuckin with everybody...
\\
\hline

\end{tabular}
\end{table}

In \textbf{FormSpring dataset}, the initial classifier's success rate was 93\% for non-hate and 75\% for hate. It misclassified 6 non-hate and 4 hate samples. Besides, after data augmentation, the accuracy improved to 98.7\%  (non-hate) and 90\% (hate).  In a total of 80 non-hate samples, only 1 sample was misclassified as a hate ( i.e., "Don't do her bad"). One possible explanation could be, this sentence contains a 'hate' part, and at the same time, it also contains a negation ("Don't"), which caused the overall verdict to be non-hate. Since negation sentences usually occur less in the dataset, the classifier might be less trained for this particular phenomenon and, thereby, fail to detect correctly. In contrast, among  the 20 hate samples,  only two were misclassified as non-hate (i.e., "love you sexy thang", and "how old are you and how do you know michael"). However, our observation revealed that authors have misunderstood these two sentences and incorrectly labeled as hate sentence. In reality, these sentences do not carry any hate, and our classifier predicted correctly.

In \textbf{Warner-Waseem dataset}, before data augmentation, our model showed a success rate of 100\% for non-hate and only 15\% for identifying hate samples. Out of 20 hate samples, it misclassified 17 samples as non-hate. However, after data augmentation, the detection of hate samples improved to 95\% and hold the previous 100\% success rate for detecting non-hate samples. In a total of 100 samples, only 1 sample was misclassified as non-hate. One possible explanation could be that this sample was comprised of multiple and long sentences (e.g., it contains 1112 characters and 232 words). Besides this sentence followed a a less common language structure (written within the African American Vernacular English), which can confuse NLP parsers, and oversampled with social slang and short-form words, e.g.,  nigga, niggazz, cuz or dis, etc, which might justify such misclassification.

In summary, from the above 300 test samples, only 35 were misclassified. Nevertheless, only 7 misclassifications occurred after dataset augmentation. Among these seven samples, one contained a negation construct, one comprised of multiple and long sentences with specific language type and social slang, spelling errors, and two were wrongly labeled. Therefore, despite the bias in the data, low ratio of hate content,  language complexity, and lack of trained dataset for specific cases, our model differentiated non-hate and hate samples 98.4\% accurately, leveraging the data augmentation method and evidence that errors might come from the dataset aspects.

Moreover, we evaluate the recall, precision and F1-score values at each class (hateful class and non-hateful class) to further distinguish classifier performance at each category. For that, we pick the dataset which presented the largest discrepancy between augmented data results and non-augmented ones and which is the Olid dataset. Table~\ref{Tab:hateClass} illustrates the results we obtained with CNN and LSTM classifiers. We can see from this table that values of recall, precision and F1-score are considerably low for the original dataset (not expanded dataset) at the hateful class compared to the non-hateful class for both classifiers. After augmentation of the dataset using our back-translation and paraphrasing methods, the values greatly increased by 35-45\% for the hateful class and by 14-20\% for the non-hateful class with CNN classification. Similarly, results of the LSTM classification increased by 28-42\% for the hateful class and by 10-19\% for the non-hateful class.

\begin{table}
\caption{Performance metrics in terms of Recall (\%), Precision (\%) and F1 score (\%) of CNN and LSTM with FastText classification representation for original and expanded Olid datasets for the hateful class.}
\label{Tab:hateClass}
\centering
\begin{tabular}{|l|l|l|l|l|l|l|l|}
\hline
\bf{} & \bf{} &\multicolumn{3}{c|}{\bf{CNN}} & \multicolumn{3}{c|}{\bf{LSTM}}  \\
\bf{Dataset} & \bf{Calculated for} & \bf{Recall} & \bf{Precision} & \bf{F1 score} & \bf{Recall} & \bf{Precision} & \bf{F1 score}\\
 \hline 

\multirow {2}{12em}{\textbf{Not expanded OLID dataset}} & \bf{Hateful class} & 48.7 & 60.0 & 53.8 & 49.6 & 67.4 & 57.1\\ 
& \bf{non Hateful class} & 83.3 & 76.0 & 79.5 & 87.7 & 77.2 & 82.1\\ \hline 
\multirow {2}{12em}{\textbf{Expanded OLID dataset}} & \bf{Hateful class} & 93.1 & 95.1 & 94.1 &  92.4 & 95.9 & 94.2\\
& \bf{non Hateful class} & 97.6 & 96.6 & 97.1 & 98.0 & 96.4 & 97.2\\ 
\hline 

\end{tabular}
\end{table}

\section{Conclusion}
\label{Sec:conclusion}
We presented in this paper a hate speech detection approach focusing on cyberbullying identification, where a new data augmentation method embedded in either a CNN or LSTM architecture for classification has been proposed and their results were compared. Before fitting data to the classifier, we expanded our datasets using a back translation approach based on a sequence to sequence architecture (Transformer architecture) and a paraphrasing technique based on mixture of experts translation model. The back translation augmentation allowed us to expand the size of the datasets to twice its initial state while the paraphrasing technique expanded our datasets almost 10 times more. Finally, the combination of both techniques helped us to expand the datasets 20 times the original size. After preprocessing, word embeddings using pretrained FastText embedding were constructed. These features are then fed to a CNN model (which later compared to LSTM model) to classify hate speech content. The performance of our methodology was evaluated on five publicly available datasets: AskFm, Formspring, Warner and Waseem, Olid and wikipedia toxic comments dataset. For each dataset, four configurations were investigated: original dataset, expansion with back-translation, expansion with paraphrasing and expansion by concatenating back-translation and paraphrasing outcomes. The results were reported in terms of accuracy, F1 score, precision and recall. We obtained a higher recall score of 99.7\% and a precision of 99.6\% for the expanded version of the Warner and Waseem dataset. Moreover, the best accuracy and F1 score of 99.4\% were recorded for the expanded Wikipedia toxic comments dataset, exceeding many state-of-the art approaches. Therefore, our findings demonstrate the efficiency and robustness of our developed approach. In the future, we desire to explore more the automatic tuning of inherent parameters of the back-translation and the paraphrasing models to enhance the quality of generated paraphrases in a way to capture the grammar and semantic meaning. In addition, we intend to explore more deep learning based classifiers and especially when considering the augmentation of the minority class (hate class) to yield better results. Again, we plan to scrutinize the generalization of the proposed methodology to other NLP related tasks instead of hate speech.

\section{Acknowledgment}
This work is supported by the European Young-sters Resilience through Serious Games, under the Internal Security Fund-Police action: 823701-ISFP-2017-AG-RAD grant, which is gratefully acknowledged.

\bibliographystyle{unsrtnat}
\bibliography{references}

\end{document}